\documentclass[11pt,twoside]{article}
\usepackage{inf_neutral}
\usepackage{epsfig}

\usepackage[dvipsnames]{xcolor}
\usepackage{hyperref}

\colorlet{mylinkcolor}{Black}
\colorlet{mycitecolor}{Black}
\colorlet{myurlcolor}{Blue}
\usepackage{graphicx}
\usepackage{cite}

\hypersetup{
  linkcolor  = mylinkcolor,
  citecolor  = black,
  urlcolor   = myurlcolor,
  colorlinks = true,
}

\begin{document}
  \title{Mammalian Value Systems}
   \author{Gopal P. Sarma \\
      School of Medicine, Emory University, Atlanta, GA USA\\
      gopal.sarma@emory.edu \\
      \AndAuthor
      Nick J. Hay \\
      Vicarious FPC, San Francisco, CA USA\\
      nnickhay@gmail.com}
  \titleodd{Mammalian Value Systems}
  \authoreven{Sarma and Hay}

\abstract{
Characterizing human values is a topic deeply interwoven with the sciences, humanities, political 
philosophy, art, and many other human endeavors.  In recent years, a number of thinkers have argued 
that accelerating trends in computer science, cognitive science, and related disciplines foreshadow the 
creation of intelligent machines which meet and ultimately surpass the cognitive abilities of human 
beings, thereby entangling an understanding of human values with future technological development.  
Contemporary research accomplishments suggest increasingly sophisticated AI systems becoming 
widespread and responsible for managing many aspects of the modern world, from preemptively 
planning users' travel schedules and logistics, to fully autonomous vehicles, to domestic robots 
assisting in daily living.  The extrapolation of these trends has been most forcefully described in the 
context of a hypothetical ``intelligence explosion,'' in which the capabilities of an intelligent software 
agent would rapidly increase due to the presence of feedback loops unavailable to biological 
organisms.  The possibility of superintelligent agents, or simply the widespread deployment of 
sophisticated, autonomous AI systems, highlights an important theoretical problem: the need to 
separate the cognitive and rational capacities of an agent from the fundamental goal structure, or value 
system, which constrains and guides the agent's actions.  The ``value alignment problem'' is to specify 
a goal structure for autonomous agents compatible with human values.  In this brief article, we suggest 
that ideas from affective neuroscience and related disciplines aimed at characterizing 
neurological and behavioral universals in the mammalian class provide important conceptual 
foundations relevant to describing human values.  We argue that the notion of ``mammalian value 
systems'' points to a potential avenue for fundamental research in AI safety and AI ethics.
}

\abstractSi{}

\maketitle

\section{Introduction}

Artificial intelligence, a term coined in the 1950's at the now famous Dartmouth Conference, has come 
to have a widespread impact on the modern world \cite{russell2003artificial, nilsson2009quest}.  If we 
broaden the phrase to include all software, and in particular, software responsible for the control and 
operation of physical machinery, planning and operations management, or other tasks requiring 
sophisticated information processing, then it goes without saying that artificial intelligence has become 
a critical part of the infrastructure supporting modern human society.  Indeed, prominent venture 
capitalist Mark Andresseen famously wrote that ``software is eating the world,'' in reference to the 
ubiquitous deployment of software systems across all industries and organizations, and the 
corresponding growth of the financial investment into software companies 
\cite{andreessen2011software}. \\

Nonetheless, there is a fundamental gap between the abilities of the most sophisticated software-based 
control systems today and the capacities of a human child or even many animals.  Our AI systems have 
yet to display the capacity for learning, creativity, independent thought and discovery that define human 
intelligence.  It is a near-consensus position, however, that at some point in the future, we will be able to 
create software-based agents whose cognitive capacities rival those of human beings.  While there is 
substantial variability in researchers' forecasts about the time-horizons of the critical breakthroughs and 
the consequences of achieving human-level artificial intelligence, there it is little disagreement that it is 
an attainable milestone \cite{muller2016future, 2017arXiv170508807G}.\footnote{There have been a number of 
prominent thinkers who have expressed strongly conservative viewpoints about AI timelines.  
See, for example, commentaries by David Deutsch, Rodney Brooks, and Douglas Hofstadter 
\cite{brooks2017, deutsch2012, hofstadter_skeptic}.}  \\

Some have argued that the creation of human-level artificial intelligence would be followed by an 
``intelligence explosion,'' whereby the intelligence of the software-based system would rapidly increase 
due to its ability to analyze, model, and improve its cognition by re-writing its codebase, in a feat of 
self-improvement impossible for biological organisms.  The net result would be a ``superintelligence,'' that 
is, an agent whose fundamental cognitive abilities vastly exceed our own 
\cite{bostrom2014superintelligence, shanahan2015technological, good1965speculations, 
chalmers2010singularity}.  \\

To be more explicit, let us consider a superintelligence to be any agent which can surpass the sum total 
of human cognitive and emotional abilities.  These abilities might include intellectual tasks such as 
mathematical or scientific research, artistic invention in musical composition or poetry, political 
philosophy and the crafting of public policy, or social skills and the ability to recognize and respond to 
human emotions.  Many commentators in recent years and decades have predicted that convergent 
advances in computer science, robotics, and related disciplines will give rise to the development of 
superintelligent machines during the 21st century \cite{muller2016future}.  \\

If it is possible to create a superintelligence, then a number of natural questions arise: What would such 
an agent choose to do?  What are the constraints that would guide its actions and to what degree can 
these actions be shaped by the designers?  If a superintelligence can reason about and influence the 
world to a substantially greater degree than human beings themselves, how can we design a system to 
be compatible with human values?  Is it even possible to formalize the notion of human values?  Are 
human values a monolithic, internally consistent entity, or are there intrinsic conflicts and contradictions 
between the values of individuals and between the value systems of different cultures?  
\cite{bostrom2014superintelligence, chalmers2010singularity, yudkowsky2008artificial, 
russell2016should, omohundro2014autonomous, omohundro2008basic}.  \\

It is our belief that the value alignment problem is of fundamental importance both for its relevance to 
near-term developments likely to be realized by the computer and robotics industries and for longer-
term possibilities of more sophisticated AI systems leading to superintelligence.  Furthermore, the 
broader set of problems posed by the realization of intelligent, autonomous, software-based agents may 
provide an important unifying framework that brings together disparate areas of inquiry spanning 
computer science, cognitive science, philosophy of mind, behavioral neuroscience, and anthropology, to 
name just a few.  \\

In this article, we set aside the question of how, when, and if AI systems will be developed that are of 
sufficient sophistication to require a solution to the value alignment problem.  This is a substantial topic 
in its own right which has been analyzed elsewhere.  We assume the feasibility of these systems as a 
starting point for further analysis of the goal structures of autonomous agents and propose the notion of 
``mammalian value systems'' as providing a framework for further research. 

\section{Goal Structures for Autonomous Agents}
\subsection{The Orthogonality Thesis}
The starting point for discussing AI goal structures is the observation that the cognitive capacities of an 
intelligent agent are independent of the goal structure that constrains or guides the agents' actions, 
what Bostrom calls the ``orthogonality thesis:''
\begin{quote}
{\small
We have seen that a superintelligence could have a great ability to shape the future according to its 
goals.  But what will its goals be?  What is the relation between intelligence and motivation in an 
artificial agent?  Here we develop two theses.  The orthogonality thesis holds (with some caveats) that 
intelligence and final goals are independent variables: any level of intelligence could be combined with 
any final goal.  The instrumental convergence thesis holds that superintelligent agents having any of a 
wide range of final goals will nevertheless pursue similar intermediary goals because they have 
common instrumental reasons to do so.  Taken together, these theses help us to think about what a 
superintelligent agent would do.\cite{bostrom2014superintelligence}
}
\end{quote}

The orthogonality thesis allows us to illustrate the importance of autonomous agents being guided by 
human-compatible goal structures, whether they are truly superintelligent as Bostrom envisions, or even 
more modestly intelligent but highly sophisticated AI systems likely to be developed in industry in the future.  
Consider the example of a domestic robot that is able to clean the house, monitor a 
security system, and prepare meals independently and without human intervention.  A robot with a 
slightly incorrect or inadequately specified goal structure might correctly infer that a household pet has 
high nutritional value to its owners, but not recognize its social and emotional relationship to the family.  
We can easily imagine the consequences for companies involved in creating domestic robots if a family 
dog or cat ends up on the dinner plate \cite{russell2016should}.  Although such a scenario is unlikely 
without some amount of warning\footnote{What exactly counts as sufficient warning, and whether 
the warning is heeded or not, is another matter.}---we may notice odd or annoying behavior in the robot in other tasks, 
for example---it highlights an important nuance about value alignment.  For example, 
the exact difference between animals that we value for their emotional role in our lives versus 
those that many have deemed ethically acceptable for food is far from obvious.  
Indeed for someone who lives on a farm, the line can be blurred and some creatures may play both roles.\\

As the intelligent capabilities of an agent grows, the consequences for slight deviations from human 
values will become greatly magnified.  The reason is that such an agent possesses increasing capacity 
to achieve its goals, however arbitrary those goals might be.  It is for this reason that researchers 
concerned with the value alignment problem have distanced themselves from the fictitious and absurd 
scenarios portrayed in Hollywood thrillers.  These movies often depict outright malevolent agents whose 
explicit aim is to destroy or enslave humanity.  What is \emph{implicit} in these stories is a goal structure 
that has been \emph{explicitly defined} to be in opposition to human values.  But as the simple example 
of the domestic robot illustrates, this is hardly the risk we face with sophisticated AI systems.  The true 
risk is that if we incorrectly or inadequately specify the goals of a sufficiently capable agent, then it will 
devote its cognitive capacities to a task that is at odds with our values in ways that may be subtle or 
even bizarre.  In the example given above, there was no malevolence or ulterior motive behind the 
robot making a nutritious meal out of the household pet.  Rather, it simply did not recognize---due to the 
failure of its human designers---that the pet was valued by its owners, not for nutritional reasons, but 
rather for social and emotional ones \cite{yudkowsky2008artificial, russell2016should}.

\subsection{Anthropomorphic Bias Versus Anthropomorphic Design}
Before proceeding, we mention an important caveat with regards to the orthogonality thesis, namely, 
that it is not a free orthogonality.  The particular goal structure of an agent will almost certainly constrain 
the necessary cognitive capabilities required for the agent to operate.  In other words, the orthogonality 
thesis does not suggest that one can pair an arbitrary set of algorithms with an 
arbitrary goal structure.  For instance, if we are building an AI system to process a large number of 
photographs and videos so that families can efficiently find their most memorable moments amidst 
terabytes of data, we know that the underlying algorithms will be those from computer vision and not 
computer algebra.  The primary takeaway from the orthogonality thesis is that when reasoning about 
intelligence in the abstract, we should not assume that any particular goal structure is implied.  In 
particular, there is no reason to believe that an arbitrary AI system having the cognitive capacity of 
humans will necessarily have a goal structure compatible with or in opposition to that of humans.  It may 
very well be completely arbitrary from the perspective of human values.  \\

This observation about the orthogonality thesis brings to light an important point with regards to AI goal 
structures, namely the difference between \emph{anthropomorphic bias} and \emph{anthropomorphic 
design}.  Anthropomorphic bias refers to the \emph{default assumption} that an arbitrary AI system will 
behave in a manner possessing commonalities with human beings.  In practice, instances of 
anthropomorphic bias almost always go hand in hand with the assumption of malevolent intentions on 
behalf of an AI system---recall our previous dismissal of Hollywood thrillers depicting agents intent on 
destroying or enslaving humanity.  \\

On the other hand, it may very well be the case, perhaps even necessary, that solving the value 
alignment problem requires us to build a \emph{specific AI system} that possesses important 
commonalities with the human mind.  This latter perspective is what we refer to as 
\emph{anthropomorphic design.}\footnote{Anthropomorphic design refers to a more narrow class of 
systems than the term ``human-compatible AI,'' which has recently come into use.   See, for example, 
\href{http://www.humancompatible.ai}{The Berkeley Center for Human-Compatible AI}.}

\subsection{Inferring Human-Compatible Value Systems}
An emerging train of thought among AI safety researchers is that a human-compatible goal structure will 
have to be inferred by the AI system itself, rather than pre-programmed by the designers.   The reason 
is that human values are rich and complex, and in addition, often contradictory and conflicting.  
Therefore, if we incorrectly specify what we think to be a safe goal structure, even slight deviations can 
be magnified and lead to detrimental consequences.  On the other hand, if an AI system begins with an 
uncertain model of human values, and then begins to learn our values by observing our behavior, then 
we can substantially reduce the risks of a misspecified goal structure.  Furthermore, just as we are 
more likely to trust mathematical calculations performed by a computer than by humans, if we build an 
AI system that we know to have greater capacity than ourselves at performing those cognitive 
operations required to infer the values of other agents by observing their behavior, then we gain the 
additional benefit of knowing that these operations will be performed with greater certainty and accuracy 
than were they to be pre-programmed by human AI researchers.  \\

There is context in contemporary research for this kind of indirect inference, such as Inverse 
Reinforcement Learning (IRL) \cite{ng2000algorithms, hadfield2016cooperative} or Bayesian Inverse 
Planning (BIP) \cite{baker2011bayesian}.  In these approaches, an agent learns the values, or utility 
function, of another agent, whether it is a human, an animal, or software system, by observing its 
behavior.  While these ideas are in their nascent stages, practical techniques have already been 
developed for designing AI systems \cite{evans2015learning, evanslearning, riedl2016using, 
riedl2016computational}.  \\

Russell summarizes the notion of indirect inference of human values by stating three principles that 
should guide the development of AI systems \cite{russell2016should}:

\begin{enumerate}
\item The machine's purpose must be to maximize the realization of human values. In particular, it has 
no purpose of its own and no innate desire to protect itself.
\item The machine must be initially uncertain about what those human values are.  The machine may 
learn more about human values as it goes along, but it may never achieve complete certainty.
\item The machine must be able to learn about human values by observing the choices that we humans 
make.
\end{enumerate}

There are almost certainly many conceptual and practical obstacles that lie ahead in designing a 
system that infers the values of human beings from observing our behavior.  In particular, human 
desires can often be masked by many layers of conflicting emotions, they can often be inconsistent, 
and the desires of one individual may outright contradict the desires of another.  In the context of a 
superintelligent agent capable of exerting substantial influence on the world (as opposed to a domestic 
robot), it is natural to ask about variations in the value systems of different cultures.  It is often assumed 
that many human conflicts on a global scale stem from conflicts in the underlying value systems of the 
respective cultures or nation states.  Is it even possible, therefore, for an AI system, no matter how 
intelligent, to arrive at a consensus goal structure that respects the desires of all people and cultures?  \\

We make two observations in response to this important set of questions.  The first is that when we say 
that cultures have conflicting values, implicit in this statement are our own limited cognitive capacities 
and ability to model the behavior and mental states of other individuals and groups.  An AI system with 
capabilities vastly greater than ourselves may quickly perceive fundamental commonalities and 
avenues for conflict resolution that we are unable to envision.  \\

To motivate this scenario, we give a highly simplified example from negotiation theory.  A method known 
as ``principled negotiation'' distinguishes between \emph{values} and \emph{positions}  
\cite{fisher1987getting}.  As an example, if two friends are deciding on a restaurant for dinner, and one 
wants Indian food and the other Italian, it may be that the first person simply likes spicy food and the 
second person wants noodles.  These preferences are the \emph{values}, spicy food and noodles, that 
the corresponding \emph{positions}, Indian and Italian, instantiate.  In this school of thought, when two 
parties are attempting to resolve a conflict, they should negotiate from values, rather than positions.  
That is, if we have some desire that is in conflict with another, we should ask ourselves---whether in the 
context of a business negotiation, family dispute, or major international conflict---what the underlying 
value is that the desire reflects.  By understanding the underlying values, we may see that there is a 
mutually satisfactory set of outcomes satisfying all parties that we failed to see initially.  In this particular 
instance, if the friends are able to state their true underlying preferences, they may recognize that Thai 
cuisine will satisfy both parties.  We mention this example from negotiation theory to raise the possibility 
that what we perceive to be fundamentally conflicting values in human society might actually be 
conflicting positions arising from distinct, but reconcilable values when viewed from the perspective of a higher 
level of intelligence.  \\

The second observation is that what we colloquially refer to as the values of a particular culture, or even 
collective human values, reflect not only innate features of the human mind, but also the development 
of human society.  In other words, to understand the underlying value system that guides human 
behavior, which would ultimately need to be modeled and inferred by an AI system, it may be helpful to 
disentangle those aspects of modern cultural values which were latent, but not explicitly evident during 
earlier periods of human history. \\

Although an agent utilizing Inverse Reinforcement Learning or Bayesian Inverse Planning will learn and 
refine its model of human values by observing our behavior, it must begin with some very rough or 
approximate initial assumptions about the nature of the values it is trying to learn.  By starting from a more
accurate initial goal structure, an agent might learn from fewer examples, thus minimizing the likelihood of
real-world actions having adverse affects.  In the remainder of 
this article, we argue that the neurological substrate common to mammals and their corresponding 
behaviors may provide a framework for characterizing the structure of the initially uncertain 
value system of an autonomous, intelligent agent.  

\subsection{Mammalian Value Systems}
\emph{\textbf{Our core thesis is the following:}} What we call human values can be informally 
decomposed into \emph{1) mammalian values, 2) human cognition, and 3) several millennia of human 
social and cultural evolution}.  This decomposition suggests that contemporary research broadly 
spanning the study of animal behavior, biological anthropology, and comparative neuroanatomy may be 
relevant to the value alignment problem, and in particular, in characterizing the initially uncertain goal 
structure which is refined through observation by an AI system.  Additionally, in analyzing the 
subsequent behavioral trajectories of intelligent, autonomous agents, we can decompose the resulting 
dynamics as being guided by mammalian values merged with AI cognition.  Aspects of contemporary 
human values which are the result of incidental historical processes---the third component of our 
decomposition above---might naturally arise in the course of the evolution of the AI system (though not 
necessarily), even though they were not directly programmed into the agent.\footnote{Many human values 
communicated to children during the course of maturation and development are the result of incidental 
historical processes.  
As an example, consider the rich set of cultural norms and social rituals surrounding food preparation.  
One does not need to have lived the entire history of a given culture to learn these norms.  
The same may be true of an AI system.}  There are many factors 
that might influence the extent to which this third component of human values continues to be 
represented in the AI system.  Examples might include whether or not these values remain meaningful 
in a world where other problems had been solved and the extent to which certain cultural values which 
were perceived to be in conflict with others could be resolved with a more fundamental understanding 
stemming from the combination of mammalian values and AI cognition.\footnote{Ethical norms can 
often vary depending on resource constraints which may also be the result of incidental historical processes.  
The norms of behavior may be different in a war zone where
individuals are fighting for survival than in an affluent society during peacetime.  If a family  
struggling to survive in a war torn country is able to escape and move to a more stable region, 
these same behaviors may no longer be necessary.  
In a similar vein, imagine an AI system that has significantly impacted global affairs by solving major problems
in food or energy production or by discovering novel insights into diplomatic strategy.  
Such an agent may find that previously necessary behaviors that have a rich human history are no longer needed.  
}\\

We want to emphasize that our claim is not that mammalian values are synonymous with human 
values.  Rather, our thesis is that there are many aspects of human values which are the result of	 
historical processes driven by human cognition.  Consequently, many structural aspects of human 
experience and human society which we colloquially refer to as ``values'' are derived entities, rather 
than features of the initial AI goal structure.  As a thought experiment, consider a scenario whereby the 
fully digitized corpus of human literature, cinema, and ongoing global developments communicated via 
the Internet are analyzed and modeled by an AI system constructed around a core mammalian goal 
structure.  In the conceptual framework that we propose, this initially mammalian structure would 
gradually come to reflect the more nuanced aspects of human society as the AI refines its model of 
human values via analysis and hypothesis generation.  We also mention that as our aim in this article is 
to focus on the structure of the initial AI motivational system and not other aspects of AI more broadly, 
we set aside the possible role human interaction and feedback may play in the subsequent 
development of the AI system's cognition and instrumental values.

\subsubsection{Neural Correlates of Values: Behavioral and Neurological Foundations}

Our thesis about mammalian values is predicated on two converging lines of evidence, one primarily 
behavioral and the other primarily neuroscientific.  Behaviorally, it is not difficult to characterize 
intuitively what human values are when viewed from the perspective of the mammalian class.  Like 
many other animals, humans are social creatures and many, if not most, of our fundamental drives 
originate from our relationships with others.  Attachment, loss, anger, territoriality, playfulness, joy, 
anxiety, and love are all deeply rooted emotions that guide our behavior and which have been 
foundational elements in the emergence of human cognition, culture, and the structure of 
society\footnote{While we have mentioned several active areas of research, there are certainly 
others that we are simply not aware of.  We apologize in advance to those scholars whose work we have not cited 
here.} \cite{horswill2008men, swanson2000cerebral, swanson2012brain, barkow1995adapted, 
dehaene2007cultural, peterson2004character, schnall2008disgust, tenenbaum2011grow, 
bowlby1980attachment, porges1995orienting, cassidy2002handbook, tomasello1999culturalorigins}.  \\

The scientific study of behavior is largely the domain of the disciplines of ethology and behaviorism.  As 
we are primarily concerned with emotions, we will focus on behavioral insights and taxonomies 
originating from the sub-community of affective neuroscience, which also aims to correlate these 
behaviors with underlying neural architecture.  More formally, Panksepp and Biven categorize the 
informal list given above into seven motivational and emotional systems that are common to mammals:  
seeking, rage, fear, lust, care, panic/grief, and play \cite{panksepp2012archaeology}.  We now give 
brief summaries of each of these systems:
\begin{enumerate}
\item SEEKING: This is the system that primarily mediates exploratory behavior and also enables the 
other systems.  The seeking system can give rise to both positive and negative emotions.  For instance, 
a mother who needs to feed her offspring will go in search of food, and the resulting maternal / child 
bonding (via the CARE system; see below) creates positive emotional reinforcement.  On the other 
hand, physical threats can generate negative emotions and prompt an animal to seek shelter and 
safety.  The behaviors corresponding to SEEKING have been broadly associated with the dopaminergic 
systems of the brain, specifically regions interconnected with the ventral tegmental area and nucleus 
accumbens.  
\item RAGE: The behaviors corresponding to rage are targeted and more narrowly focused than those 
governed by the seeking system.  Rage compels animals towards specific threats and is generally 
accompanied by negative emotions.  However, it should be noted that in an adversarial scenario where 
rage can lead to victory, it can also be accompanied by the positive emotions of triumph or glory.  The 
RAGE system involves medial regions of the amygdala, medial regions of the hypothalamus, and the 
periaqueductal gray.
\item FEAR: The two systems described thus far are directly linked to externally directed, action-oriented behavior.  
In contrast, fear describes a system which places an animal in a negative affective 
state, one which it would prefer not to be in.  In the early stages, fear tends to correspond to stationary 
states, after which it can transition to seeking or rage, and ultimately, attempts to flee from the offending 
stimulus.  However, these are secondary effects, and the primary physical state of fear is typically 
considered to be an immobile one.  The FEAR system involves central regions of the amygdala, 
anterior and medial regions of the hypothalamus, and dorsal regions of the periaqueductal gray. 
\item LUST: Lust describes the system leading to behaviors of courtship and reproduction.  Like fear, it 
will tend to trigger the seeking system, but can also lead to negative affective states if satisfaction is not 
achieved. The LUST system involves anterior and ventromedial regions of the hypothalamus.
\item CARE: Care refers to acts of tenderness directed towards loved ones, and in particular, an 
animal's offspring.  As we described in the context of seeking, the feelings associated with caring and 
nurturing can be profoundly positive and play a crucial component in the social behavior of mammals.  
CARE is associated with the ventromedial hypothalamus and the oxytocin system.
\item PANIC / GRIEF: Activation of the panic / grief system corresponds to profound psychological pain, 
and is generally not associated with external physical causes.  In young animals, this system is typically 
activated by separation from caregivers, and is the underlying network behind ``separation anxiety.''  
Like care, the panic / grief system is a fundamental component of mammalian 
social behavior.  It is the negative affective system which drives animals towards relationships with 
other animals, thereby stimulating the care system, generating feelings of love and affection, and giving 
rise to social bonding.  This system is associated with the periaqueductal gray, ventral septal area, and 
anterior cingulate.
\item PLAY: The play system corresponds to lighthearted behavior in younger animals and is a key 
component of social bonding, friendship, as well as the learning of survival-oriented skills.  Although 
play can superficially resemble aggression, there are fundamental differences between play and adult 
aggression.  At an emotional level, it goes without saying that play corresponds to positive affective 
states, and unlike aggressive behavior, is typically part of a larger, orchestrated sequence of events.  In 
play, for example, animals often alternate between assuming dominant and submissive roles.  The 
PLAY system is currently less neuroanatomically localized, but involves midline thalamic regions.
\end{enumerate}

As we stated earlier, our thesis about mammalian values originates from two convergent lines of 
evidence, one behavioral and the other neuroscientific.  What we refer to as the ``neural correlates of 
values,'' or NCV, are the common mammalian neural structures which underlie the motivational and 
emotional systems summarized above.  To the extent that human values are intertwined with our 
emotions, these architectural commonalities suggest that the shared mammalian neurological substrate 
is of importance to understanding human value alignment in sophisticated learning systems.  Panksepp 
and Biven write, 

\begin{quote}
{\small To the best of our knowledge, the basic biological values of all mammalian brains were built 
upon the same basic plan, laid out in \ldots affective circuits that are concentrated in subcortical 
regions, far below the neocortical ``thinking cap'' that is so highly developed in humans.  Mental life 
would be impossible without this foundation. There, among the ancestral brain networks that we share 
with other mammals, a few ounces of brain tissue constitute the bedrock of our emotional lives, 
generating the many primal ways in which we can feel emotionally good or bad within ourselves. As we 
mature and learn about ourselves, and the world in which we live, these systems provide a solid
foundation for further mental developments \cite{panksepp2012archaeology}.}
\end{quote}

Latent in this excerpt is the decomposition that we have suggested earlier.  The separation of the 
mammalian brain into subcortical and neocortical regions, roughly corresponding to emotions and 
cognition respectively, implies that we can attempt to reason by analogy what the architecture of an AI 
system would look like with a human-compatible value system.  In particular, the initially uncertain goal 
structure that the AI system refines via observation may be much simpler than we might imagine by 
reflecting on the complexities of human society and individual desires.  As we have illustrated using our 
simple example from negotiation theory, our intuitive understanding of human values, and the conflicts 
that we regularly witness between individuals and groups, may in fact represent conflicting positions 
stemming from a shared fundamental value system, a value system that originates from the subcortical 
regions of the brain, and which other mammals share with us.\footnote{There is a contemporary and 
light-hearted social phenomenon which provides an evocative illustration of the universality of 
mammalian emotions, namely, the volume of animal videos posted to YouTube.  From ordinary citizens 
with pets, to clips from nature documentaries, animal videos are regularly watched by millions of 
viewers worldwide.  Individual videos and compilations of ``animal odd couples,'' ``unlikely animal 
friends,'' ``dogs and babies,'' and ``animal friendship between different species'' are commonly 
searched enough to be auto-completed by YouTube's search capabilities.  It is hardly surprising that 
these charming and heart-warming videos are so compelling to viewers of all age groups, genders, and 
ethnic backgrounds.  Our relationships with other animals, whether home owners and their pets, or 
scientists and the wild animals that they study, tell us something deeply fundamental about ourselves 
\cite{nytimes_mammal}.  The strong emotional bonds that humans form with other animals, in 
particular, with our direct relatives in the mammalian class, and the draw to simply watching this 
social behavior in other mammals, is a vivid illustration of the fundamental role that emotions play in our 
inner life and in guiding our behavior.  \\

In the future, the potential to apply inverse reinforcement learning (or related techniques) to large 
datasets of videos, including 
short clips from YouTube, movies, TV shows, documentaries, etc.\ opens up an interesting avenue to 
evaluate and further refine the hypothesis presented here.  For instance, when such technology 
becomes available, we might imagine comparing the inferred goal structures when restricted to videos 
of human behavior versus those restricted to mammalian behavior.  There are many other variations 
along these lines, for instance, restricting to videos of non-mammalian behavior, mammals as well as 
humans, different cultures, etc.}  \\

Referring once again to the work of Panksepp, 
\begin{quote}
{\small
In short, many of the ancient, evolutionarily derived brain systems all mammals share still serve as the 
foundations for the deeply experienced affective proclivities of the human mind.  Such ancient brain 
functions evolved long before the emergence of the human neocortex with its vast cognitive skills.  
Among living species, there is certainly more evolutionary divergence in higher cortical abilities than in 
subcortical ones \cite{panksepp1998affective}.}
\end{quote}

The emphasis on the diversity in higher cortical abilities is of particular relevance to the decomposition 
that we have proposed.  We might ask what the full spectrum of higher cortical 
abilities are that could be built on top of the common mammalian substrate provided by the 
evolutionarily older parts of the brain.  We need not confine ourselves to those manifestations of higher 
cognition that we see in nature, or that would even be hypothetical consequences of continued 
evolution by natural selection.  Indeed, one restatement of our core thesis is to consider---in the 
abstract or as a thought experiment---the consequences of extending the diversity of brain architectures 
to include higher cortical abilities arising not from natural selection, but rather the \emph{de novo} 
architectures of artificial intelligence.  

\subsection{Relationship to Moral Philosophy}
It is hardly a surprise that a vibrant area of research within AI safety is the relationship of contemporary 
and historical theories of moral philosophy to the problem of value alignment.  Indeed, researchers have 
specifically argued for the relevance of moral philosophy in the context of the inverse reinforcement 
learning paradigm (IRL) that is the starting point for analysis in this article \cite{armstrong_leike}. \\

Is the framework we propose in opposition to those that are oriented towards moral philosophy?  On the 
one hand, our perspective is that the field of AI safety is simply too young to make such judgments. At 
our present level of understanding, we believe each of these agendas form solid foundations for further 
research and there seems little reason to pursue one to the exclusion of the other.  On the other hand, we 
would also argue that this distinction is a false dichotomy.  Indeed, there are 
active areas of research in the ethics community aimed at understanding the neurological and cognitive 
underpinning of human moral reasoning \cite{greene2002and,greene2009cognitive}.  Therefore, it is 
quite possible that a hybrid approach to value alignment emerges, bridging the ``value 
primitives'' perspective we advocate here with research from moral philosophy.\footnote{In a recent article,
Baum has argued that the normative basis for ``social choice'' and ``bottom-up'' approaches to AI ethics must 
overcome strong obstacles that have been insufficiently explored by the AI safety community \cite{baum2017social}.  
Although the approach we describe here decomposes values into 
more fundamental components, it is not \emph{a priori} in opposition to top-down ethics.  In an extreme case,
one could certainly imagine employing a purely predetermined approach to ethics within the context of mammalian values in which no value learning takes place.  However, as we stated above, 
we suspect that an intermediate ground will be found when the issues are more thoroughly examined, and 
for that reason, we are reluctant to endorse either a bottom-up or a top-down approach too strongly.  
Given the intellectual youth of the field of AI safety, we see little reason to give strong preference 
to one set of approaches over the other.  Moreover, an important observation that Baum makes in 
framing his argument is that considerable work relevant to AI ethics already exists in the social choice 
literature, and yet none of this work has been discussed in any detail by the AI safety community.  
In our minds, this is a more fundamental point, namely, that there is substantial scholarship in 
many areas of academic research  relevant to AI safety.  For this reason, we believe that where there is controversy,
the first step should be to ensure that the best possible representations of given viewpoints
have been made visible and adequately discussed before endorsing particular courses of action.}

\section{Discussion}
	 
The possibility of autonomous, software-based agents, whether self-driving cars, domestic robots, or 
the longer-term possibilities of superintelligence, highlights an important theoretical problem---the need 
to separate the intelligent capabilities of such a system from the fundamental values which guide the 
agents' actions.  For such an agent to exist in a human world and to act in a manner compatible with 
human values, these values would need to be explicitly modeled and formalized.  An emerging train of 
thought in AI safety research is that this modeling process would need to be conducted by the AI 
system itself, rather than by the system's designers.  In other words, the agent would start off with an 
initially uncertain goal structure and infer human values over time by observing our behavior.  \\

The question that motivates this article is to ask the following: what can we say about the broad 
features of the initial goal structure that the agent then refines through observation and hypothesis 
generation?  The perspective we advocate is to view human values within the context of the broader 
mammalian class, thereby providing implicit priors on the latent structure of the values we aim to 
infer.  The shared neurological structures underlying mammalian emotions and their corresponding 
social behaviors provide a starting point for formalizing an initial value system for autonomous, 
software-based agents.  There are several practical implications of having a more detailed understanding
of the structure of human values.  By having more detailed prior information, it may be possible to learn from fewer examples.
For an agent that is actively making decisions and having an impact on the world, learning an ethical framework
more efficiently can minimize potential catastrophes.  Furthermore, an informative prior may make approaches to
AI safety which are otherwise computationally intractable into practical options.  \\

From this vantage point, we argue that what we colloquially refer to as human values can be informally 
decomposed into \emph{1) mammalian values,  2) human cognition, and 3) several millennia of human 
social and cultural evolution}.  In the context of a \emph{de novo} artificially intelligent agent, we can 
characterize desirable, human-compatible behavior as being described by mammalian values merged 
with AI cognition.  It goes without saying that we have left out a considerable amount of detail in this 
description.  The specifics of Inverse Reinforcement Learning, the many neuroscientific nuances 
underlying the comparative neuroanatomy, physiology, and function of the mammalian brain, as well as 
the controversies and competing theories in the respective disciplines are all substantial topics on their 
own right.  \\

Our omission of these issues is not out of lack of recognition or belief that they are unimportant.  Rather, 
our aim in this article has been to present a high-level overview of a richly interdisciplinary 
set of questions whose broad outlines have only recently begun to take shape.  We will tackle these 
issues and others in a subsequent series of manuscripts and invite interested researchers to join us.  
Our fundamental motivation in proposing this framework is to bring together scholars from diverse 
communities that may not be aware of each other's research and their potential for synergy.  We believe 
that there is a wealth of existing research which can be fruitfully re-examined and re-conceptualized 
from the perspective of artificial intelligence and the value alignment problem.  We hope that additional 
interaction between these communities will help to refine and more precisely define research problems 
relevant to designing safe AI goal structures. 
\subsection*{Acknowledgements}
We would like to thank Adam Safron, Owain Evans, Daniel Dewey, Miles Brundage, 
and several anonymous reviewers for insightful discussions and feedback 
on the manuscript.  We would also like to thank the guest editors of \emph{Informatica}, 
Ryan Carey, Matthijs Maas, Nell Watson, and Roman Yampolskiy, for organizing this special issue.

\bibliographystyle{ieeetr}
\bibliography{mammalian_values}

\end{document}